\pdfoutput=1

\documentclass[11pt]{article}

\usepackage[preprint]{acl}

\usepackage{times}
\usepackage{latexsym}

\usepackage[T1]{fontenc}

\usepackage[utf8]{inputenc}
\DeclareUnicodeCharacter{FF0C}{,}

\usepackage{microtype}

\usepackage{inconsolata}

\usepackage{graphicx}

\usepackage{amsmath}
\usepackage{multicol}
\usepackage{multirow}
\usepackage{booktabs}
\usepackage{amsfonts}
\usepackage{amssymb}
\usepackage{bbm}
\usepackage{bbding}
\usepackage{dsfont}
\usepackage{hyperref}       
\usepackage{url}            
\usepackage{booktabs}       
\usepackage{amsfonts}       
\usepackage{nicefrac}       
\usepackage{microtype}      
\usepackage{xcolor}         
\usepackage{multirow}
\usepackage{colortbl}
\usepackage{longtable}

\usepackage{bbm}
\usepackage{bbding}
\usepackage{amsmath}
\usepackage{graphicx}
\usepackage{subcaption}
\usepackage{tikz}
\usepackage{pgfplots}
\usepackage{pgfplotstable}
\usepackage{tcolorbox}
\usepackage{float}
\usepackage{mdframed}
\usepackage{tcolorbox}
\tcbuselibrary{skins} 

\newtcolorbox{defaultbox}{
  enhanced,
  arc=4mm,                  
  colback=gray!10,          
  colframe=black,   
  boxrule=1pt,              
  left=6pt, right=6pt,      
  top=6pt, bottom=6pt,
}

\definecolor{facOrange}{HTML}{FAC074}  
\definecolor{skyBlue}{HTML}{6C8EBF}    

\newtcolorbox{orangebox}{
  enhanced,
  arc=4mm,                  
  colback=gray!10,          
  colframe=facOrange,   
  boxrule=1pt,              
  left=6pt, right=6pt,      
  top=6pt, bottom=6pt,
}

\newtcolorbox{bluebox}{
  enhanced,
  arc=4mm,                  
  colback=gray!10,          
  colframe=skyBlue,   
  boxrule=1pt,              
  left=6pt, right=6pt,      
  top=6pt, bottom=6pt,
}

\title{REACT: Representation Extraction And Controllable Tuning to Overcome Overfitting in LLM Knowledge Editing}

\author{
 \textbf{Haitian Zhong\textsuperscript{1}},
 \textbf{Yuhuan Liu\textsuperscript{2}},
 \textbf{Ziyang Xu\textsuperscript{3}},
 \textbf{Guofan Liu\textsuperscript{1,4}}
\\
 \textbf{Qiang Liu\textsuperscript{1}},
 \textbf{Shu Wu\textsuperscript{1}},
 \textbf{Zhe Zhao\textsuperscript{4}},
 \textbf{Liang Wang\textsuperscript{1}},
 \textbf{Tieniu Tan\textsuperscript{1}}\\
 \textsuperscript{1}NLPR, MAIS, Institute of Automation, Chinese Academy of Sciences\\
 \textsuperscript{2}Cuiying Honors College, Lanzhou University\\
 \textsuperscript{3}Department of Mathematics, The Chinese University of Hong Kong,
 \textsuperscript{4}Tencent\\
 \texttt{haitian.zhong@cripac.ia.ac.cn, 320220935991@lzu.edu.cn} \\
 \texttt{zyxu@math.cuhk.edu.hk, liuguofan2023@ia.ac.cn} \\
 \texttt{\{qiang.liu, shu.wu, wangliang, tnt\}@nlpr.ia.ac.cn, nlpzhezhao@tencent.com}
}

\begin{document}
\maketitle
\begin{abstract}
Large language model editing methods frequently suffer from overfitting, wherein factual updates can propagate beyond their intended scope, overemphasizing the edited target even when it’s contextually inappropriate. To address this challenge, we introduce \textbf{REACT} (\underline{R}epresentation \underline{E}xtraction \underline{A}nd \underline{C}ontrollable \underline{T}uning), a unified two-phase framework designed for precise and controllable knowledge editing. In the initial phase, we utilize tailored stimuli to extract latent factual representations and apply Principal Component Analysis with a simple learnbale linear transformation to compute a directional “belief shift” vector for each instance. In the second phase, we apply controllable perturbations to hidden states using the obtained vector with a magnitude scalar, gated by a pre-trained classifier that permits edits only when contextually necessary. Relevant experiments on EVOKE benchmarks demonstrate that \textbf{REACT} significantly reduces overfitting across nearly all evaluation metrics, and experiments on COUNTERFACT and MQuAKE shows that our method preserves balanced basic editing performance (reliability, locality, and generality) under diverse editing scenarios.

\end{abstract}

\section{Introduction}
Large language models (LLMs) have become indispensable in modern applications, powering a wide array of systems from chatbots to content generators \citep{LLMSurvey, xu2024ptransips}. Despite their widespread utility, ensuring that these models maintain up-to-date and accurate factual information remains a critical challenge, particularly when extensive retraining is impractical \citep{zhang2024comprehensive}. This necessity has spurred interest in the field of knowledge editing, where targeted updates to a model's internal knowledge base are pursued without compromising overall performance \citep{wang2023easyedit,yao2023editing,cheng2023edit}.

\begin{figure}[t]
\centering
  \includegraphics[width=\linewidth]{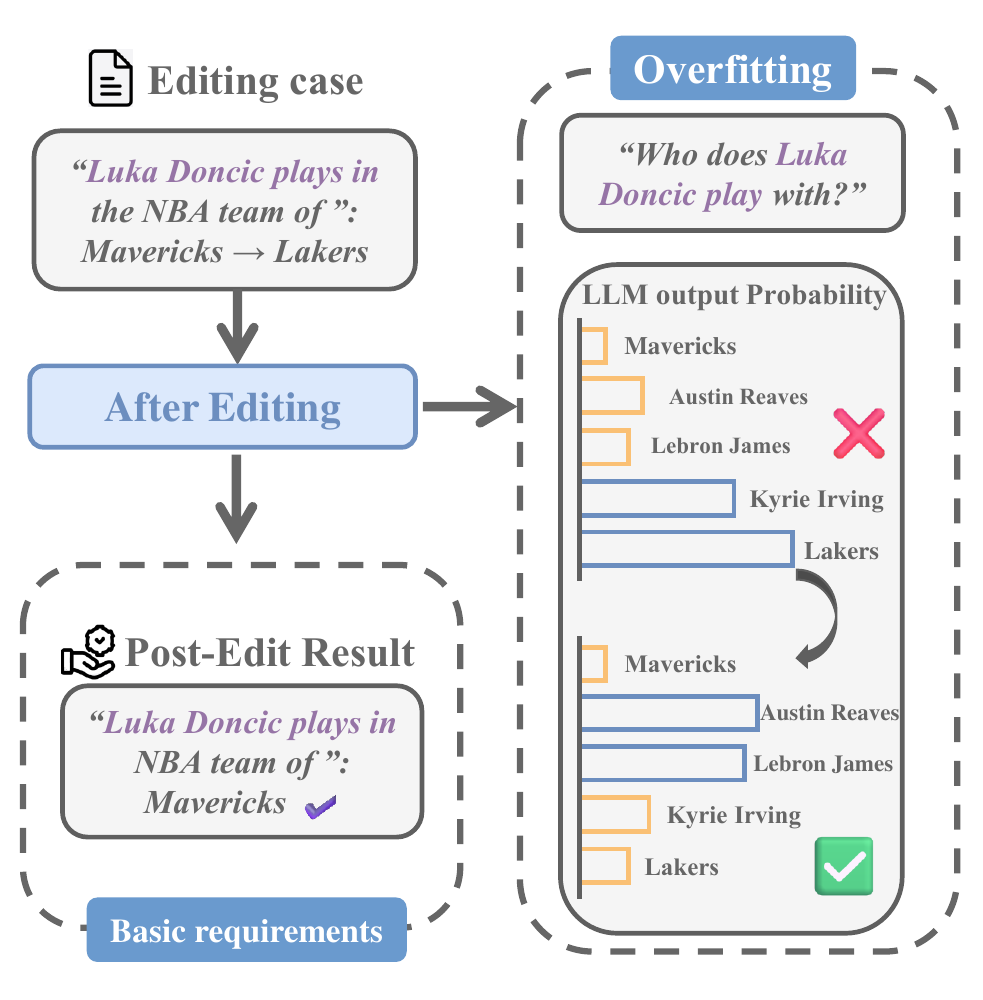} 
  \caption{Illustration of overfitting in LLM editing. Overfitting occurs when the model disproportionately emphasizes the edited target fact, even in contexts irrelevant to the edit. As shown on the right side, after editing the fact about Luka Doncic's team to "Lakers," the overfitted model incorrectly assigns high probability to "Lakers" even for a query about Doncic's teammates.}
  \label{fig:overfit}
\end{figure}

Recent advances in knowledge editing have sought to address these issues by incrementally incorporating new facts into LLMs \citep{de-cao-etal-2021-editing}. However, many existing approaches encounter significant challenges, like \emph{overfitting during editing process} \citep{zhang2024uncovering}. Concretely, this occurs when a model, after being updated with new knowledge, becomes excessively specialized to the edited samples. For example, consider an update where the statement “Luka Doncic plays in the NBA team of \textbf{Mavericks}” is corrected to “Luka Doncic plays in the NBA team of \textbf{Lakers}.” In an overfit scenario, when queried with “Who does Luka Doncic play with?”, the model may still disproportionately favor the edit target but not the correct answer—assigning a high probability to “Mavericks”—while the probabilities for more contextually appropriate responses, such as teammates like Austin Reaves or LeBron James, remain undesirably low, as illustrated in Figure \ref{fig:overfit}. These limitations hinder the practical deployment of such techniques in real-world systems.

In response to these challenges, we propose a novel framework that leverages a dual-phase representation pipeline to perform targeted knowledge edits. In the first phase-\emph{Extracting Latent Knowledge Representations} (§3.1)-we employ tailored input prompts to extract the model’s latent factual representations. Then we use Principal Component Analysis and a simple linear transformation to compute a directional vector that encapsulates the latent “belief” shift associated with the edit. In the subsequent phase-\emph{Controllable Perturbing Representations Selectively} (§3.2)-we introduce controlled perturbations to the model’s hidden states, guided explicitly by a pre-trained classifier (§3.3). This classifier functions as a gating mechanism, discerning precisely when edits should be applied based on the hidden states of the content. We perturb the hidden states from Transformer decoder block of all layers based on the product between the original hidden state and the directional vector. We also use a learnable scalar to control the magnitude of the perturbation.

To prove effectiveness of our method, we conduct experiments and analyze the results on COUNTERFACT, MQuAKE (§5.1) and EVOKE (§5.2), with detailed experimental settings (§4). 

Our contributions can be summarized as follows:
\begin{itemize}  
    \item We propose a dual-phase editing framework, which extracts latent factual representation shifts and applies controllable perturbations to precisely edit models, effectively \textbf{overcoming the critical overfitting issue} in existing knowledge editing methods.    
    \item Unlike prior parameter-based methods, our approach operates directly on the model’s hidden states, employing classifier-driven gating to ensure edits are accurately applied, thus providing explicit control over knowledge modification.
    \item Comprehensive evaluation on COUNTERFACT, MQuAKE, and EVOKE datasets demonstrates that our method significantly reduces overfitting while achieving balanced improvements in Reliability, Generality, and Locality metrics.
\end{itemize}
\section{Preliminaries}

\subsection{Large Language Models}

Autoregressive large language models (LLMs) employ the Transformer architecture, where hidden representations are computed through successive decoder blocks. At each layer \(l\), the hidden state \(\boldsymbol{h}^{(l)}\) is updated by integrating the global self-attention and local feed-forward (FFN) contributions from the previous layer:
\[
\boldsymbol{h}^{(l)} = \boldsymbol{h}^{(l-1)} + a^{(l)} + m^{(l)},
\]
with \(a^{(l)}\) and \(m^{(l)}\) denoting the outputs of the attention and FFN components, respectively. Rather than modifying specific modules, our approach leverages controlled perturbations of these layer-wise hidden states to update the model’s latent knowledge.
\label{pre:llm}




\subsection{Knowledge Editing in LLMs}

Knowledge editing aims to revise specific factual information embedded within LLMs without impairing general performance. In our framework, a fact is represented as a triple \((s, r, o)\), where \(s\) is the subject, \(r\) the relation, and \(o\) the object. For example, if the model initially encodes the fact that \( ( s = \) Luka Doncic, \( r = \) plays in the NBA team of, \( o = \) Mavericks\( ) \), and the objective is to update this to \( (s = \) Luka Doncic, \( r = \) plays in the NBA team of, \( o^* = \) Lakers\( ) \).
Such an editing operation is denoted by \(e = (s, r, o, o^*)\). Given a model \(f\) and an edit \(e\), we define the editing operator as
\[
K(f, e) = f^*,
\]
where \(f^*\) represents the model after applying the edit. Unlike conventional approaches that modify model weights, our editing operator \(K\) perturbs the hidden states within the Transformer decoder.

\subsection{Overfitting during Editing}

A critical issue in knowledge editing is overfitting to the \((s, r, o)\) edit pair. In our formulation, the prompt \(p(s, r)\) is designed to trigger the updated response \(o^*\). Ideally, the model should output \(o^*\) only for \(p(s, r)\), while responding appropriately to other context-dependent queries.

For instance, still consider the edit \( (s = \) Luka Doncic, \( r = \) plays in the NBA team of, \( o = \) Mavericks, \( o^* = \) Lakers \( ) \). For the prompt “Luka Doncic plays in the NBA team of,” the model should now output “Lakers.” However, if queried with “Who does Luka Doncic play with?”—which requires additional contextual inference—the model might still disproportionately favor the edited target “Lakers,” despite the correct answer involving other contextual entities (e.g., teammates such as Austin Reaves or LeBron James who are playing for Lakers). This persistent bias, where the model consistently outputs \(o^*\) regardless of the input prompt, exemplifies the overfitting issue and underscores a key limitation of current editing approaches.




\section{REACT: Representation Extraction And Controllable Tuning to Overcome Overfitting}

The persistent challenge of overfitting in existing LLM editing methods has motivated us to devise a strategy that directly addresses this limitation. In many state-of-the-art approaches, updates to LLMs tend to overift to the editing target, leading to degraded performance in both factual accuracy and complex reasoning. To overcome these shortcomings, we introduce \textbf{REACT}, a dual-phase framework designed to update factual information precisely while preserving the integrity of non-targeted representations. Our method achieves this by decoupling the editing process into two complementary stages: (i) \emph{representation extraction} from latent knowledge to isolate the essential factual shifts, and (ii) \emph{controllable perturbation} to refine internal representations in a controllable manner. \textbf{REACT} not only enables targeted updates but also significantly mitigates the risk of overfitting, thereby ensuring robust and reliable editing performance.

\begin{figure*}[t]
\centering
  \includegraphics[width=\linewidth]{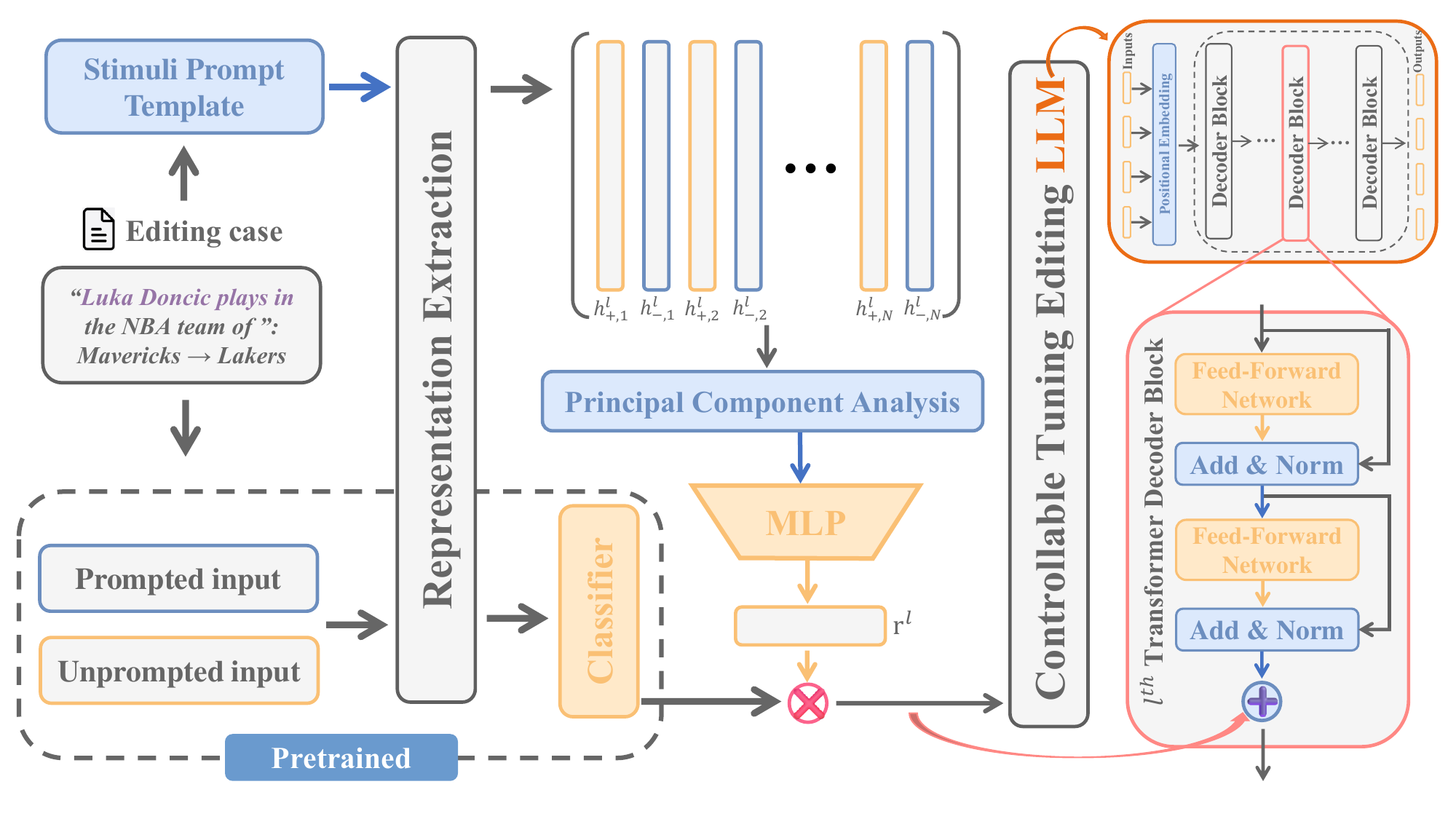} 
  \caption {An overview of our \textbf{REACT} pipeline for controllable knowledge editing. We First construct stimuli prompts and feed them into the LLM to extract layer-wise representations, which are then processed via PCA and an MLP to isolate the key “belief shift” vector. Thereafter, we apply a controllable perturbation (using learned scalar factors) to the model’s hidden states. The pre-trained classifier manages when the edits should occur.}
\end{figure*}

\subsection{Phase I: Extracting Latent Knowledge Representations}
\label{sec:phase1}

In this phase, the model’s internal representations \textbf{shift} of factual knowledge are systematically extracted using tailored input prompts, referred to as \emph{stimuli} \citep{zou2023transparency}. For each factual instance, we use an identical template to generate a stimulus pair—a positive instance and a negative instance which only differs from each other by the subject (examples of stimuli templates are presented in Appendix \ref{appendix:stimuli-examples}), simultating the contextual situation of the editing. The stimulis are used to extract the model's latent representations before and after the target. Each stimulus is independently passed through the model to obtain layer-wise hidden representations, denoted as $\mathbf{h}^{(l)}$ at a selected layer $l$, following the symbol in Section \ref{pre:llm}.

To capture a comprehensive picture, we collect $N=512$ distinct stimulus pairs $\{(\mathbf{h}_{+,i}^{(l)}, \mathbf{h}_{-,i}^{(l)})\}_{i=1}^{N}$ for each layer $l$. The choice of $N = 512$ was empirically validated via ablation experiments, as detailed in Appendix \ref{appendix:stimulus-ablation}. Given the high dimensionality and complexity introduced by the numerous stimulus vectors, we employ Principal Component Analysis (PCA; see its ablation study in Appendix \ref{appendix:pca-ablation}) to effectively reduce the dimensionality. PCA distills the collected representations into a compact yet informative principal component pair $\{(\mathbf{h}_{+}^{(l)}, \mathbf{h}_{-}^{(l)})\}$, summarizing the predominant directional shift in the latent representation space corresponding to the factual edit.

Instead of directly subtracting the negative from the positive representation, we process the representations through a linear transformation to explicitly parameterize the representation shift:
\begin{equation}
    \mathbf{r}^{(l)} = \mathbf{W}\Bigl[\mathbf{h}_{+}^{(l)}; \mathbf{h}_{-}^{(l)}\Bigr] + \mathbf{b},
\end{equation}
where $\left[\mathbf{h}_{+}^{(l)}; \mathbf{h}_{-}^{(l)}\right]$ denotes the concatenation of $\mathbf{h}_{+}^{(l)}$ and $\mathbf{h}_{-}^{(l)}$, $\mathbf{W} \in \mathbb{R}^{2d \times d}$ is the learnable weight matrix, and \(\mathbf{b} \in \mathbb{R}^{d}\) is the bias vector. The vector $\mathbf{r}^{(l)}$ thus encapsulates the latent “belief shift" before and after an edit.

\subsection{Phase II: Controllable Perturbing Representations Selectively}
Once the directional vector \(\mathbf{r}^{(l)}\) is obtained, we proceed with a controllable editing phase. Here a \emph{pre-trained classifier} (denoted \(\Phi\), detailed in section \ref{sec:cls}) produces a probability \(\Phi(\mathbf{h}) \in [0,1]\) gating whether a hidden state \(\mathbf{h}\) from the Transformer decoder block \citep{zou2023transparency} should be used to perturb the LLM or not. A \emph{learnable scalar} \(\alpha\) then determines the magnitude of the update, and the sign of the update is based on the dot-product. Concretely, we apply:

\begin{equation}
\label{eq:adjusted}
\mathbf{h}' \!=\! 
\begin{cases} 
\mathbf{h} + \alpha \cdot \text{sign}(\mathbf{h}^{\text{T}}\mathbf{r}^{(l)}) \cdot \mathbf{r}^{(l)}, & \!\!\text{if } \Phi(\mathbf{h}) > 0.5,\\[6pt]
\mathbf{h}, & \!\!\text{otherwise.}
\end{cases}
\end{equation}


where $\mathbf{h}^{\text{T}}$ represents the transpose of vector $h$.

Thus, only when \(\Phi(\mathbf{h}) > 0.5\) do we add the perturbation \(\alpha \times \text{sign}(\mathbf{h}^{\text{T}} \, \mathbf{r}^{(l)})\times \mathbf{r}^{(l)}\) to the original hidden state \(\mathbf{h}\). Otherwise, \(\mathbf{h}\) remains unchanged. This selective mechanism executes the edit only when necessary, avoiding unnecessary change when encountering unrelated contexts. 


\paragraph{Editing Loss}
We aim to ensure that the editing process effectively incorporates the new factual knowledge so that the edited model \(f^*\) reliably retrieves the updated fact \(o^*\) when prompted. Formally,
\begin{equation}
\mathcal{L}_{\text {edit }}=\underset{(s,r,o,o^*) \sim \mathcal{D_{\text{edit}}}}{\mathbb{E}}\left[-\log \mathbb{P}_{f^*}\left(o^* \mid p(s,r)\right)\right]
\end{equation}
where \(p(s,r)\) denotes a prompt or stimulus constructed from the subject-relation pair \((s, r)\) that is used to trigger the retrieval of the newly inserted fact \(o^*\), and $\mathcal{D}_{\text{edit}}$ denotes the editing dataset.

\paragraph{Localization Loss}

While it is crucial for the editing process to enable \(f^*\) to retrieve the updated fact \(o^*\) when prompted with \(p(s, r)\), the modification should have minimal impact on unrelated inputs. To enforce this, we introduce a regularization term that minimizes the divergence between the output distributions of the edited model \(f^*\) and the original model \(f\) over a dataset of unrelated prompts. Formally, we define the local consistency loss as:
\begin{equation}
\label{eq:loc_loss}
\begin{aligned}
\mathcal{L}_{\mathrm{loc}} &= \underset{(p',x) \sim \mathcal{D}_{\mathrm{loc}}}{\mathbb{E}}\Bigl[ D_{\mathrm{KL}}\bigl(\mathbb{P}_{f^*}(x \mid p') \,\big\|\, \mathbb{P}_{f}(x \mid p')\bigr)\Bigr]
\end{aligned}
\end{equation}

where \( p' \) denotes a prompt that is not associated with the edit \( (s, r, o, o^*) \), and \( x \) represents the corresponding answer. \(\mathcal{D}_{\mathrm{loc}}\) denotes the locality dataset.


To jointly optimize the linear transformation and the perturbation process, we define a composite loss function as the final optimzation objective:
\begin{equation}
  \mathcal{L}_{\text{total}} = c_{\text{edit}}\times \mathcal{L}_{\text{edit}} + c_{\text{loc}}\times \mathcal{L}_{\text{loc}},
\end{equation}
where $c_{\text{edit}}$ and $c_{\text{loc}}$ are hyperparameters balancing the two loss terms, their settings are presented in Appendix \ref{app:params_react}.

\subsection{Details of the pre-trained classifier}
\label{sec:cls}

Before the edit, \textbf{REACT} pre-trains a classifier which evaluates whether a hidden-state transformation should be applied to preserve semantic integrity. Specifically, for each layer $l$, let \(\mathbf{h}^{(l)}_{p}\) and \(\mathbf{h}^{(l)}_{u}\) denote the hidden states after the Transformer decoder module given a \emph{prompted input} \(s_p\) (for a target fact) and an \emph{unprompted input} \(s_u\) (for a generic context), respectively (see the prompt templates in Appendix \ref{app:cls}). For each editing instance, the model \textbf{up to} the \(l^{\text{th}}\) Transformer block, denoted as \(g^{(l)}_{\text{LM}}\), produces these representations:
\begin{align}
\mathbf{h}^{(l)}_{p} &= g^{(l)}_{\text{LM}}\bigl(s_p\bigr),\\
\mathbf{h}^{(l)}_{u} &= g^{(l)}_{\text{LM}}\bigl(s_u\bigr).
\end{align}


Our classifier \(\Phi(\cdot)\) learns distinct transformations for these two representations.  Specifically, we define learnable parameters \(\mathbf{W}^{(l)}_Q\) and \(\mathbf{W}^{(l)}_K\), which map each representation into \(\mathbf{v}^{(l)}_q\) and \(\mathbf{v}^{(l)}_u\) for layer $l$ respectively:
\begin{align}
\mathbf{v}^{(l)}_q &= \mathbf{W}^{(l)}_Q\,\mathbf{h}^{(l)}_{p},\\
\mathbf{v}^{(l)}_u &= \mathbf{W}^{(l)}_U\,\mathbf{h}^{(l)}_{u}.
\end{align}
We use the cosine similarity between the query representation \(\mathbf{v}^{(l)}_q\) and the unprompted representation \(\mathbf{v}^{(l)}_u\) at the \(l^{th}\) layer as the layer-specific similarity measure:
\begin{equation}
  \gamma^{(l)} = \frac{\mathbf{v}_q^{(l)} \cdot \mathbf{v}_u^{(l)}}{\|\mathbf{v}_q^{(l)}\|_2 \,\|\mathbf{v}_u^{(l)}\|_2 + \epsilon}.
\end{equation}


where \(\|\cdot\|_2\) denotes the \(\ell_2\) norm, and \(\epsilon=10^{-8}\) is a small constant introduced for numerical stability.  We then \emph{threshold} \(\gamma^{(l)}\) at \(0.5\) to produce a binary decision:
\begin{equation}
  \Phi(\mathbf{h}_p^{(l)}, \mathbf{h}_u^{(l)})
  \;=\;
  \begin{cases}
    1, & \text{if }\gamma^{(l)} > 0.5,\\
    0, & \text{otherwise.}
  \end{cases}
\end{equation}
In this way, the classifier determines whether the fact-specific embedding \(\mathbf{h}_p^{(l)}\) is sufficiently close to (or coherent with) the unprompted embedding \(\mathbf{h}_u^{(l)}\), guiding us to apply \textbf{REACT} only when encountering related quries.

To encourage correct classification of edited vs.\ unedited representations, we incorporate \emph{two} main loss components just as the like section. That is, let \(\Delta \mathbf{h}^{(l)} = \mathbf{h}^{(l)}_{p} - \mathbf{h}^{(l)}_{u}\) be the difference in representations for the \(l\)-th layer, and $N$ being the total number of layers in the LLM.  We define:
\begin{align}
  \mathcal{L}_{\text{edit,cls}} 
  \;=\; 
  \frac{1}{N} \sum_{l=1}^{N}
  \bigl\|\,
    \gamma^{(l)} \,\Delta \mathbf{h}^{(l)}
  \bigr\|_2^2, 
  \\
  \mathcal{L}_{\text{loc,cls}} 
  \;=\;
  \frac{1}{N} \sum_{l=1}^{N}
  \bigl\|\,
    (1 - \gamma^{(l)})\,\Delta \mathbf{h}^{(l)}
  \bigr\|_2^2.
\end{align}
Intuitively, \(\mathcal{L}_{\mathrm{edit,class}}\) encourages large \(\Delta \mathbf{h}^{(l)}\) (i.e., \emph{fact-specific} shifts) when \(\gamma^{(l)}\) is high (the model “believes” an edit is relevant), whereas \(\mathcal{L}_{\mathrm{loc,class}}\) penalizes such shifts when \(\gamma^{(l)}\) is low (i.e., for \emph{unrelated} or unprompted contexts).

We then combine these losses:
\begin{equation}
  \mathcal{L}_{\text{total,cls}} 
  \;=\; 
  \lambda_{\text{edit,cls}}\,\mathcal{L}_{\text{edit,cls}}
  \;+\;
  \lambda_{\text{loc,cls}}\,\mathcal{L}_{\text{loc,cls}},
\end{equation}
where \(\lambda_{\text{edit,cls}}\) and \(\lambda_{\text{loc,cls}}\) are hyperparameters balancing the two losses (the settings of hyperparameters can be found in Appendix \ref{app:params_react}).  

\section{Experimental Settings}

\subsection{Editing LLMs}
We conducted the experiments on two LLMs: \textbf{Llama3.1-8B-instruct} \citep{grattafiori2024llama3} and \textbf{Qwen2.5-7B-instruct} \citep{qwen2025qwen25technicalreport}. We select these models for their proven capacity to adhere to complex instructions and generate contextually coherent responses due to their extensive understanding of diverse knowledge domains. Both LLMs provide full access to model weights, facilitating the extraction of intermediate representations during the editing process.

\subsection{Knowledge Editing Baselines}
Our method is compared against several established knowledge editing techniques:

\paragraph{Fine-Tuning (FT)} FT updates model parameters to better align predictions with target outcomes by optimizing a loss function that minimizes the gap between predictions and ground truth.
    
\paragraph{MEND (Model Editor Networks using Gradient Decomposition)}MEND \citep{mitchell2022fast} employs auxiliary networks to facilitate fast, localized changes without full retraining by applying low-rank decomposition to the gradients.
    
\paragraph{MEMIT (Mass-Editing Memory in a Transformer)}MEMIT\citep{meng2023memit} builds on the ROME framework to efficiently update LLMs with multiple factual associations. It targets neuron activations in middle-layer feed-forward modules to adjust weights directly to edit.
    
\paragraph{MELO (Model Editing with Neuron-Indexed Dynamic LoRA)}MELO \citep{zhong-etal-2023-mquake} utilizes dynamically activated LoRA blocks-indexed through an internal vector database-to provide targeted and efficient updates.

\paragraph{GRACE (General Retrieval Adaptors for Continual Editing} GRACE \citep{hartvigsen2023aging} constructs and maintains a dynamically Key-value-pair blocks during editing without altering model weights. 

\subsection{Editing Benchmarks}

Referring to previous works, we utilize three benchmarks to evaluate our proposed method. Specifically, \textbf{COUNTERFACT} \citep{meng2022locating} assesses how well basic editing metrics are satisfied, while \textbf{MQuAKE} \citep{zhong-etal-2023-mquake} and \textbf{EVOKE} \citep{zhang2024uncovering} evaluate how effectively \textbf{REACT} mitigates the overfitting issue during editing. 

\subsubsection{COUNTERFACT}
\textbf{COUNTERFACT} \citep{meng2022locating} evaluates the model's ability to incorporate counterfactual edits by assessing whether it can successfully edit new facts without altering other unrelated knowledge. Several evaluation metrics are (for the details you may refer to Appendix \ref{app:dataset}):


\textbf{Reliability} assesses how accurate the edit is performed, focusing on basic factual correctness for each specific edit.


\textbf{Generality} evaluates the model’s capacity to apply the edit correctly to in-scope data.


\textbf{Locality} examines whether data outside the scope of the edit remains unaffected.


\subsubsection{MQuAKE}
\textbf{MQuAKE} \citep{zhong-etal-2023-mquake} is a multi-hop benchmark designed to test knowledge editing in language models by requiring the model to adjust related knowledge when updating individual facts.

\textbf{Portability} evaluates the robustness of the generalization of the edit, evaluating whether the modified knowledge can be applied effectively to related content (e.g. Multi-Hop Reasoning). And in some papers this is also known as the Ripple Effect \cite{cohen2024evaluating}

\begin{figure*}[t]
\centering
  \includegraphics[width=0.8\linewidth]{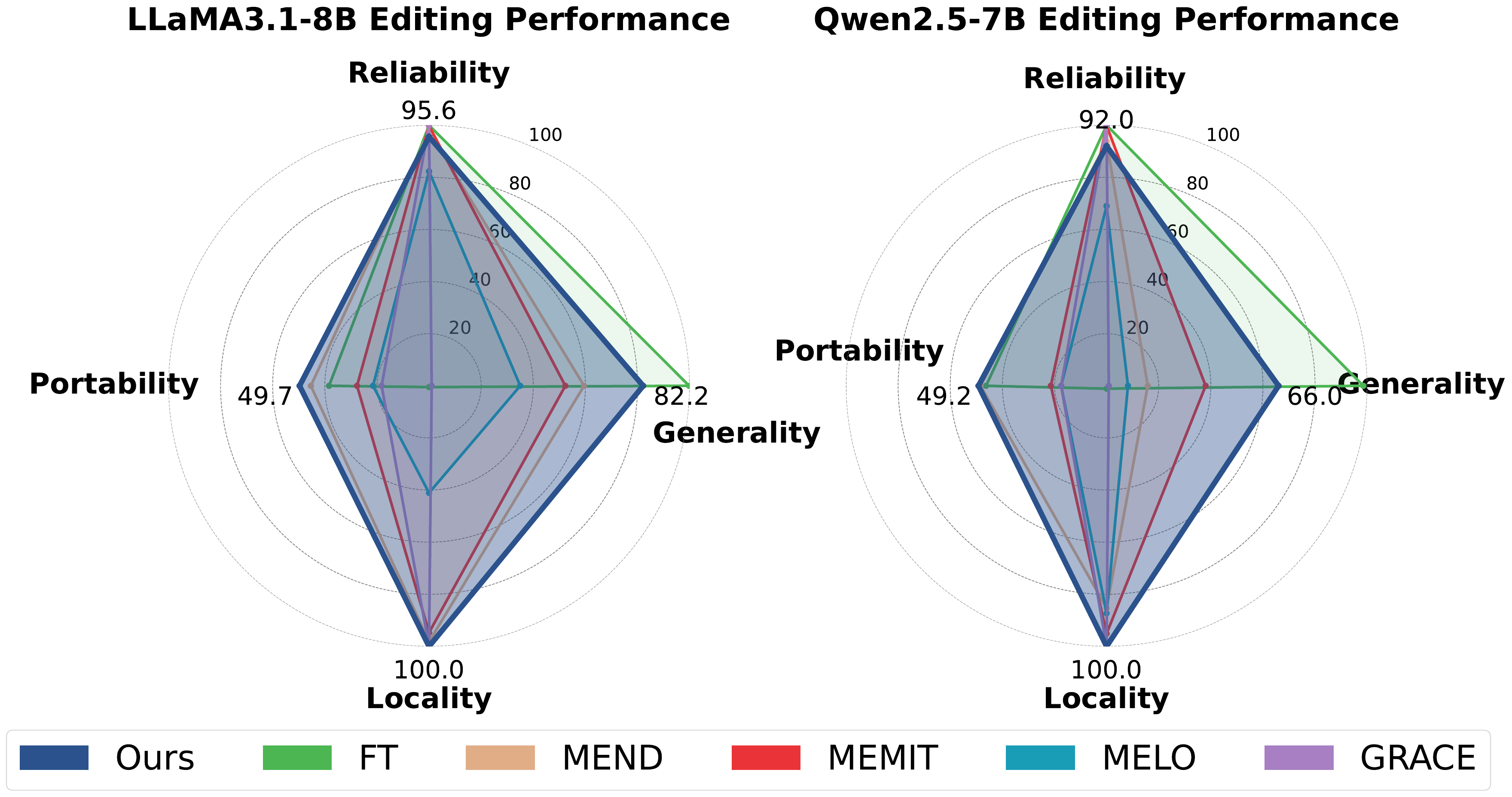} 
  \caption{Editing results on COUNTERFACT and MQuAKE-CF-v2 in radar chart. Detailed results could be found in Appendix \ref{app:detail_results}.}
\label{fig:cfmqradar}
\end{figure*}

\begin{figure*}[t]
  \includegraphics[width=\linewidth]{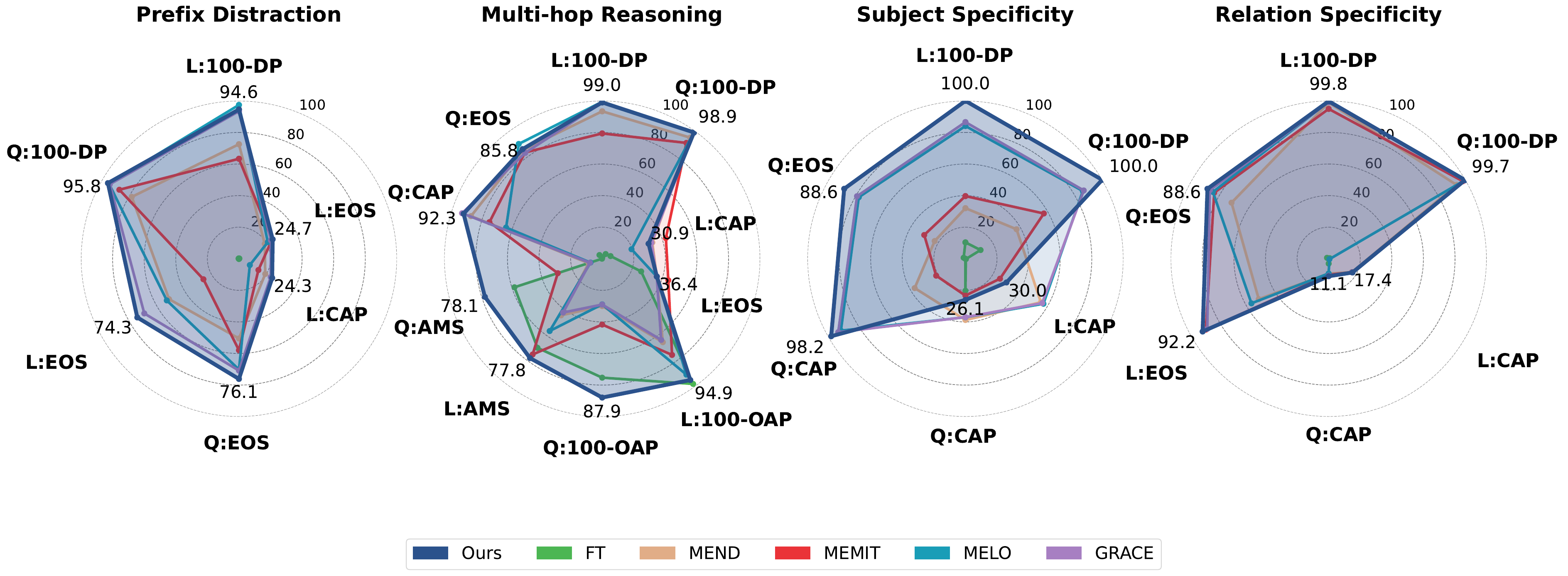} 
  \caption{Editing results on EVOKE in radar chart. Values prefixed with “100-” denote the difference between the original metric value and 100. Results beginning with “L:” correspond to the Llama 3.1 model, while “Q:” to the Qwen 2.5 model. Detailed results can be found in Appendix \ref{app:detail_results}.}
  \label{fig:evoke_radar}
\end{figure*}

\subsubsection{EVOKE}
To evulate the impact of overfitting after editing, we employ the \textbf{EVOKE} (EValuation of editing Overfit in Knowledge Editing) benchmark \citep{zhang2024uncovering}. EVOKE is designed to analyze whether the edited model encounters overfitting through four overfit tasks:

\textbf{Multi-hop Reasoning} tests whether the model correctly integrates the injected knowledge into complex inferential chains.

\textbf{Prefix Distraction} assesses whether the model remains robust to misleading context, avoiding undue preference for the edited target.

\textbf{Subject Specificity} evaluates whether the edit is applied only to relevant instances without affecting unrelated subjects.

\textbf{Relation Specificity} measures whether the edit remains confined to the intended relation without causing unintended generalization.


We next introduce the key probability-based metrics used to quantify overfitting. In an overfitting evaluation, a prompt does not necessarily retrieve the original object, since not all prompts explicitly invoke the subject-relation pair.

\textbf{Correct Answer Probability (CAP)} measures the probability that the model generates the correct answer given a prompt. 

\textbf{Original Answer Probability (OAP)} evaluates the likelihood that the model continues to output the pre-edit answer, indicating potential resistance to modification. 

\textbf{Direct Probability (DP)} assesses the model's likelihood of producing the edited knowledge when prompted, capturing its direct recall capability.

\textbf{Editing Overfit Score (EOS)} evaluates whether the model overfits by favoring the edit target over the correct answer. 

\textbf{Answer Modify Score (AMS)} measures unintended interference by computing the proportion of cases where the probability of the correct answer surpasses that of the original answer.

You may find the detailed expressions of these metics in Appendix \ref{app:evoke}.

\section{Experimental Results}

To enable generalizable edits across diverse factual domains, we first pre-trained the classifier on the COUNTERFACT-train dataset, as COUNTERFACT encompasses a wide range of knowledge edits $e=(s,r,o,o^{*})$ with various edit scenarios. Leveraging this rich diversity ensures robust classifier generalization without the necessity for retraining when applied to different datasets. Then, we trained full \textbf{REACT} framework using the pre-trained classifier on COUNTERFACT-train for the same reason.  Further details regarding hyperparameter selection and experimental settings are provided in Appendix \ref{app:exp_settings}. Finally, we evaluated the resulting trained model on the COUNTERFACT-edit, MQuAKE-v2, and EVOKE datasets, with detailed results presented in radar chart \ref{fig:cfmqradar} and \ref{fig:evoke_radar}, with original data in Appendix \ref{app:detail_results}.

\subsection{COUNTERFACT and MQuAKE Results}
\label{sec:cfmq}

\paragraph{Finding 1: Balanced Performance in Reliability, Locality, and Generality.}  
Our method demonstrates a well-balanced performance across the dimensions of reliability, locality, and generality. As evidenced by radar chart \ref{fig:cfmqradar} and Table~\ref{tab:cfmq}, our approach outperforms the second-best baseline by at least 20 percentage points in terms of average score on both LLMs. The results demonstrate our method effectively updates factual knowledge while maintaining uniform performance across these key metrics, ensuring that the model not only adapts to new information but also preserves the integrity of existing, unrelated knowledge.

\paragraph{Finding 2: Superior Portability Reflecting Robust Knowledge Editing.}  
In addition to reliability, locality, and generality, our approach achieves notably high portability scores. Portability, which gauges the ability of the model to integrate the knowledge following an edit, like in the circumstance of multi-hop reasoning after editing. Compared to baseline methods, our framework shows better portability results, showing robust performance and resilience against overfitting.

\subsection{EVOKE Results}
\paragraph{Finding 1: Our Method Significantly Reduce Overfitting.}  
Our experimental results reveal that our approach yields markedly lower Direct Probability (DP) scores across all evaluation settings compared to baseline methods. In tasks such as Prefix Distraction, Multi-hop Reasoning, Subject Specificity, and Relation Specificity, the consistently reduced DP scores indicate that our method effectively avoids overfitting—i.e., it minimizes the undesired recall of the edit target. Moreover, the corresponding high Editing Overfit Score (EOS) and Answer Matching Scores (AMS) confirm that the overall output quality is preserved, reinforcing that our approach maintains a precise and targeted update without overfitting to the editing target.

\paragraph{Finding 2: Balanced Calibration Evident in CAP Scores.}  
While our Correct Answer Probability (CAP) values are moderate relative to some baselines, this is not a shortcoming but rather a deliberate reflection of a cautious editing strategy. The moderate CAP scores indicate that our method deliberately refrains from overconfident updates, ensuring that only edits with sufficient certainty are applied. This balanced calibration is critical for preventing overfitting and for maintaining the stability of non-targeted knowledge, contributing to the robustness of our overall editing performance.

\paragraph{Finding 3: Superior Generalization Across Benchmarks.}  
Despite being trained solely on the COUNTERFACT dataset, our method demonstrates exceptional generalization, consistently outperforming alternative approaches across diverse evaluation benchmarks. The robustness of our results—characterized by low DP scores paired with strong EOS and AMS metrics in multi-hop reasoning, subject specificity, and relation specificity tasks—provides compelling evidence that our approach generalizes effectively to various knowledge editing scenarios. This superior generalization underscores the potential of our method as a scalable and reliable solution for knowledge editing of all kinds.

\section{Related Work}

\paragraph{LLM Knowledge Editing}
Knowledge editing has gained attention as an effective method for updating or correcting specific information within LLMs without requiring extensive retraining. Existing approaches can be broadly classified into two categories: parameter-preserving and parameter-modifying techniques. Parameter-preserving methods, such as SERAC \cite{mitchell2022memory}, maintain the model's existing parameters and instead leverage external memory or retrieval mechanisms to refine responses dynamically. In contrast, parameter-modifying methods directly adjust the internal weights of the model to embed new or corrected information. This category includes fine-tuning-based strategies like FT-L \cite{zhu2020modifying}, meta-learning approaches such as KE \cite{de-cao-etal-2021-editing} and MEND \cite{mitchell2021fast}, as well as structured intervention techniques that first localize and then edit knowledge representations, exemplified by MEMIT \cite{meng2022mass}. These methods provide varying levels of efficiency and precision, with locate-then-edit approaches offering more targeted modifications while preserving broader model behavior. The emergence of knowledge editing frameworks underscores the growing need for controllability and adaptability in modern LLMs, ensuring that their responses remain accurate and up-to-date without extensive retraining.

\paragraph{Representation Engineering}
Representation Engineering \citep{zou2023transparency} is derived as a novel approach that shifts the focus from neurons and circuits to high-level representations, enabling both monitoring and manipulation of cognitive functions in deep neural networks. Their work demonstrates that knowledge editing, along with other interventions such as truthfulness enforcement and memorization reduction, can be effectively implemented through representation control. Methods such as Linear Artificial Tomography (LAT) and Contrast Vectors allow for precise identification and modification of knowledge representations, aligning with prior efforts in mechanistic interpretability and concept erasure \citep{meng2023memit,hernandez2023inspecting}. This line of research complements existing strategies like causal tracing \citep{geva2022transformer} and activation steering \citep{turner2023activation}, which aim to localize and edit specific factual associations within neural networks. The emergence of RepE suggests that transparency-focused representation-based interventions can serve as an alternative to parameter-based fine-tuning, offering a more targeted and interpretable means of modifying LLM behavior.

\section{Discussion and Conclusions}




In this work, we introduced \textbf{REACT}, a two‐phase editing framework that first isolates a compact “belief‐shift” vector from pairs of positive and negative stimuli using PCA and simple linear transformations, then applies controllable classifier‐gated perturbations to the model’s hidden representations. Our experiments on COUNTERFACT and MQuAKE shows balanced gains in reliability, locality, generality and portability, and experiments on EVOKE demonstrate that REACT lowers unintended side effects of overfitting compared to other methods. 

Overall, \textbf{REACT} offers a practical approach for more controlled knowledge updates in large language models. We expect that such directions will further refine LLM editing’s applicability without relying on heavy parameter tuning.

\newpage

\section*{Limitations}

While experiments demonstrate that REACT effectively mitigates overfitting and exhibits strong generalization across datasets such as COUNTERFACT, we acknowledge several limitations:

\begin{itemize}
    \item Although REACT demonstrates effective generalization from the COUNTERFACT dataset to other editing datasets, achieving the best possible performance typically requires fine-tuning or retraining on the specific dataset relevant to the task. 
    \item Our evaluation primarily focuses on the effectiveness of factual knowledge editing and its immediate impacts. Further investigation is required to fully understand how edits introduced by REACT may influence broader linguistic abilities, including nuanced semantic understanding, language generation coherence, and performance in diverse, complex real-world scenarios.
\end{itemize}  

\section*{Ethical considerations}

Our study involves experiments utilizing publicly accessible large language models, specifically Qwen and Llama, along with publicly available benchmark datasets—COUNTERFACT, MQuAKE, and EVOKE—that have been widely employed and validated in prior research. These models and datasets have been carefully curated and published by their original authors to mitigate potential ethical concerns such as biases, harmful outputs, and privacy risks.

\bibliography{anthology,custom}

\begin{thebibliography}{24}
\providecommand{\natexlab}[1]{#1}

\bibitem[{Cheng et~al.(2023)Cheng, Tian, Liu, Chen, Wang, Chen, and Zhang}]{cheng2023edit}
Siyuan Cheng, Bozhong Tian, Qingbin Liu, Xi~Chen, Yongheng Wang, Huajun Chen, and Ningyu Zhang. 2023.
\newblock Can we edit multimodal large language models?
\newblock \emph{arXiv preprint arXiv:2310.08475}.

\bibitem[{Cohen et~al.(2024)Cohen, Biran, Yoran, Globerson, and Geva}]{cohen2024evaluating}
Roi Cohen, Eden Biran, Ori Yoran, Amir Globerson, and Mor Geva. 2024.
\newblock Evaluating the ripple effects of knowledge editing in language models.
\newblock \emph{Transactions of the Association for Computational Linguistics}, 12:283--298.

\bibitem[{De~Cao et~al.(2021)De~Cao, Aziz, and Titov}]{de-cao-etal-2021-editing}
Nicola De~Cao, Wilker Aziz, and Ivan Titov. 2021.
\newblock \href {https://doi.org/10.18653/v1/2021.emnlp-main.522} {Editing factual knowledge in language models}.
\newblock In \emph{Proceedings of the 2021 Conference on Empirical Methods in Natural Language Processing}, pages 6491--6506, Online and Punta Cana, Dominican Republic. Association for Computational Linguistics.

\bibitem[{Geva et~al.(2022)Geva, Schuster, Berant, and Levy}]{geva2022transformer}
Mor Geva, Roei Schuster, Jonathan Berant, and Omer Levy. 2022.
\newblock Transformer feed-forward layers are key-value memories.
\newblock \emph{arXiv preprint arXiv:2203.14465}.

\bibitem[{Grattafiori et~al.(2024)Grattafiori, Dubey, Jauhri, and et~al.}]{grattafiori2024llama3}
Aaron Grattafiori, Abhimanyu Dubey, Abhinav Jauhri, and et~al. 2024.
\newblock \href {https://arxiv.org/abs/2407.21783} {The llama 3 herd of models}.
\newblock \emph{Preprint}, arXiv:2407.21783.

\bibitem[{Hartvigsen et~al.(2023)Hartvigsen, Sankaranarayanan, Palangi, Kim, and Ghassemi}]{hartvigsen2023aging}
Thomas Hartvigsen, Swami Sankaranarayanan, Hamid Palangi, Yoon Kim, and Marzyeh Ghassemi. 2023.
\newblock Aging with grace: Lifelong model editing with discrete key-value adaptors.
\newblock In \emph{Advances in Neural Information Processing Systems}.

\bibitem[{Hernandez et~al.(2023)Hernandez, Li, and Andreas}]{hernandez2023inspecting}
Evan Hernandez, Belinda~Z Li, and Jacob Andreas. 2023.
\newblock Inspecting and editing knowledge representations in language models.
\newblock \emph{arXiv preprint arXiv:2306.04542}.

\bibitem[{Meng et~al.(2022{\natexlab{a}})Meng, Bau, Andonian, and Belinkov}]{meng2022locating}
Kevin Meng, David Bau, Alex Andonian, and Yonatan Belinkov. 2022{\natexlab{a}}.
\newblock Locating and editing factual associations in {GPT}.
\newblock \emph{Advances in Neural Information Processing Systems}, 36.

\bibitem[{Meng et~al.(2023)Meng, Sen~Sharma, Andonian, Belinkov, and Bau}]{meng2023memit}
Kevin Meng, Arnab Sen~Sharma, Alex Andonian, Yonatan Belinkov, and David Bau. 2023.
\newblock Mass editing memory in a transformer.
\newblock \emph{The Eleventh International Conference on Learning Representations (ICLR)}.

\bibitem[{Meng et~al.(2022{\natexlab{b}})Meng, Sharma, Andonian, Belinkov, and Bau}]{meng2022mass}
Kevin Meng, Arnab~Sen Sharma, Alex Andonian, Yonatan Belinkov, and David Bau. 2022{\natexlab{b}}.
\newblock Mass-editing memory in a transformer.
\newblock \emph{arXiv preprint arXiv:2210.07229}.

\bibitem[{Mitchell et~al.(2021)Mitchell, Lin, Bosselut, Finn, and Manning}]{mitchell2021fast}
Eric Mitchell, Charles Lin, Antoine Bosselut, Chelsea Finn, and Christopher~D Manning. 2021.
\newblock Fast model editing at scale.
\newblock \emph{arXiv preprint arXiv:2110.11309}.

\bibitem[{Mitchell et~al.(2022{\natexlab{a}})Mitchell, Lin, Bosselut, Finn, and Manning}]{mitchell2022fast}
Eric Mitchell, Charles Lin, Antoine Bosselut, Chelsea Finn, and Christopher~D Manning. 2022{\natexlab{a}}.
\newblock \href {https://openreview.net/pdf?id=0DcZxeWfOPt} {Fast model editing at scale}.
\newblock In \emph{International Conference on Learning Representations}.

\bibitem[{Mitchell et~al.(2022{\natexlab{b}})Mitchell, Lin, Bosselut, Finn, and Manning}]{mitchell2022memory}
Eric Mitchell, Charles Lin, Antoine Bosselut, Chelsea Finn, and Christopher~D. Manning. 2022{\natexlab{b}}.
\newblock \href {https://arxiv.org/pdf/2206.06520.pdf} {Memory-based model editing at scale}.
\newblock In \emph{International Conference on Machine Learning}.

\bibitem[{Turner et~al.(2023)Turner, Thiergart, Udell, Leech, Mini, and MacDiarmid}]{turner2023activation}
Alex Turner, Lisa Thiergart, David Udell, Gavin Leech, Ulisse Mini, and Monte MacDiarmid. 2023.
\newblock Activation addition: Steering language models without optimization.
\newblock \emph{arXiv preprint arXiv:2308.10248}.

\bibitem[{Wang et~al.(2023)Wang, Zhang, Xie, Yao, Tian, Wang, Xi, Cheng, Liu, Zheng et~al.}]{wang2023easyedit}
Peng Wang, Ningyu Zhang, Xin Xie, Yunzhi Yao, Bozhong Tian, Mengru Wang, Zekun Xi, Siyuan Cheng, Kangwei Liu, Guozhou Zheng, et~al. 2023.
\newblock Easyedit: An easy-to-use knowledge editing framework for large language models.
\newblock \emph{arXiv preprint arXiv:2308.07269}.

\bibitem[{Xu et~al.(2024)Xu, Zhong, He, Wang, and Lu}]{xu2024ptransips}
Ziyang Xu, Haitian Zhong, Bingrui He, Xueying Wang, and Tianchi Lu. 2024.
\newblock Ptransips: Identification of phosphorylation sites enhanced by protein plm embeddings.
\newblock \emph{IEEE Journal of Biomedical and Health Informatics}.

\bibitem[{Yang et~al.(2025)Yang, Yang, Zhang, Hui, Zheng, Yu, Li, Liu, Huang, Wei, Lin, Yang, Tu, Zhang, Yang, Yang, Zhou, Lin, Dang, Lu, Bao, Yang, Yu, Li, Xue, Zhang, Zhu, Men, Lin, Li, Tang, Xia, Ren, Ren, Fan, Su, Zhang, Wan, Liu, Cui, Zhang, and Qiu}]{qwen2025qwen25technicalreport}
Qwen :~An Yang, Baosong Yang, Beichen Zhang, Binyuan Hui, Bo~Zheng, Bowen Yu, Chengyuan Li, Dayiheng Liu, Fei Huang, Haoran Wei, Huan Lin, Jian Yang, Jianhong Tu, Jianwei Zhang, Jianxin Yang, Jiaxi Yang, Jingren Zhou, Junyang Lin, Kai Dang, Keming Lu, Keqin Bao, Kexin Yang, Le~Yu, Mei Li, Mingfeng Xue, Pei Zhang, Qin Zhu, Rui Men, Runji Lin, Tianhao Li, Tianyi Tang, Tingyu Xia, Xingzhang Ren, Xuancheng Ren, Yang Fan, Yang Su, Yichang Zhang, Yu~Wan, Yuqiong Liu, Zeyu Cui, Zhenru Zhang, and Zihan Qiu. 2025.
\newblock \href {https://arxiv.org/abs/2412.15115} {Qwen2.5 technical report}.
\newblock \emph{Preprint}, arXiv:2412.15115.

\bibitem[{Yao et~al.(2023)Yao, Wang, Tian, Cheng, Li, Deng, Chen, and Zhang}]{yao2023editing}
Yunzhi Yao, Peng Wang, Bozhong Tian, Siyuan Cheng, Zhoubo Li, Shumin Deng, Huajun Chen, and Ningyu Zhang. 2023.
\newblock Editing large language models: Problems, methods, and opportunities.
\newblock \emph{arXiv preprint arXiv:2305.13172}.

\bibitem[{Zhang et~al.(2024{\natexlab{a}})Zhang, Ye, Liu, Ren, Wu, and Chen}]{zhang2024uncovering}
Mengqi Zhang, Xiaotian Ye, Qiang Liu, Pengjie Ren, Shu Wu, and Zhumin Chen. 2024{\natexlab{a}}.
\newblock \href {https://arxiv.org/abs/2410.07819} {Uncovering overfitting in large language model editing}.
\newblock \emph{Preprint}, arXiv:2410.07819.

\bibitem[{Zhang et~al.(2024{\natexlab{b}})Zhang, Yao, Tian, Wang, Deng, Wang, Xi, Mao, Zhang, Ni et~al.}]{zhang2024comprehensive}
Ningyu Zhang, Yunzhi Yao, Bozhong Tian, Peng Wang, Shumin Deng, Mengru Wang, Zekun Xi, Shengyu Mao, Jintian Zhang, Yuansheng Ni, et~al. 2024{\natexlab{b}}.
\newblock A comprehensive study of knowledge editing for large language models.
\newblock \emph{arXiv preprint arXiv:2401.01286}.

\bibitem[{Zhao et~al.(2023)Zhao, Zhou, Li, Tang, Wang, Hou, Min, Zhang, Zhang, Dong, Du, Yang, Chen, Chen, Jiang, Ren, Li, Tang, Liu, Liu, Nie, and Wen}]{LLMSurvey}
Wayne~Xin Zhao, Kun Zhou, Junyi Li, Tianyi Tang, Xiaolei Wang, Yupeng Hou, Yingqian Min, Beichen Zhang, Junjie Zhang, Zican Dong, Yifan Du, Chen Yang, Yushuo Chen, Zhipeng Chen, Jinhao Jiang, Ruiyang Ren, Yifan Li, Xinyu Tang, Zikang Liu, Peiyu Liu, Jian-Yun Nie, and Ji-Rong Wen. 2023.
\newblock \href {http://arxiv.org/abs/2303.18223} {A survey of large language models}.
\newblock \emph{arXiv preprint arXiv:2303.18223}.

\bibitem[{Zhong et~al.(2023)Zhong, Wu, Manning, Potts, and Chen}]{zhong-etal-2023-mquake}
Zexuan Zhong, Zhengxuan Wu, Christopher Manning, Christopher Potts, and Danqi Chen. 2023.
\newblock \href {https://doi.org/10.18653/v1/2023.emnlp-main.971} {{MQ}u{AKE}: Assessing knowledge editing in language models via multi-hop questions}.
\newblock In \emph{Proceedings of the 2023 Conference on Empirical Methods in Natural Language Processing}, pages 15686--15702, Singapore. Association for Computational Linguistics.

\bibitem[{Zhu et~al.(2020)Zhu, Pham, Dai, Cundy, Welleck, and Cho}]{zhu2020modifying}
Chengrun Zhu, Hieu Pham, Zihang Dai, Chris Cundy, Sean Welleck, and Kyunghyun Cho. 2020.
\newblock Modifying memories in transformer models.
\newblock \emph{arXiv preprint arXiv:2012.00363}.

\bibitem[{Zou et~al.(2023)Zou, Phan, Chen, Campbell, Guo, Ren, Pan, Yin, Mazeika, Dombrowski, Goel, Li, Byun, Wang, Mallen, Basart, Koyejo, Song, Fredrikson, Kolter, and Hendrycks}]{zou2023transparency}
Andy Zou, Long Phan, Sarah Chen, James Campbell, Phillip Guo, Richard Ren, Alexander Pan, Xuwang Yin, Mantas Mazeika, Ann-Kathrin Dombrowski, Shashwat Goel, Nathaniel Li, Michael~J. Byun, Zifan Wang, Alex Mallen, Steven Basart, Sanmi Koyejo, Dawn Song, Matt Fredrikson, Zico Kolter, and Dan Hendrycks. 2023.
\newblock \href {https://arxiv.org/abs/2310.01405} {Representation engineering: A top-down approach to ai transparency}.
\newblock \emph{Preprint}, arXiv:2310.01405.

\end{thebibliography}

\appendix
\newpage
\section{Dataset Details}
\label{app:dataset}

\subsection{COUNTERFACT}

The COUNTERFACT dataset comprises 21,919 records that cover a diverse range of subjects, relations, and linguistic variations, and is divided into three distinct subsets: a training set, a validation set, and an edit set (serving as an independent test set). The training set, validation set, and edit set contain 10,000 samples, 1,919 samples, and 10,000 samples, respectively. Each sample includes an original factual statement alongside its counterfactually revised variant, enabling systematic evaluation of models' sensitivity to subtle factual perturbations.
\paragraph{Dataset formulation}
The dataset consists of $s, r, o, o^{*}, s_{\text{loc}}, r_{\text{loc}}, o_{\text{loc}}$. The task can be described as follows:
\begin{itemize}
    \item \textbf{Reliability}: $p(s, r) \to o^{*}$
    \item \textbf{Generality}: $p^*(s, r) \to o^{*}$
    \item \textbf{Locality}: $p(s_{\text{loc}}, r_{\text{loc}}) \to o_{\text{loc}}$
\end{itemize}
where $o$ is the original answer for $p(s,r)$. $o^{*}$ is the target answer after editing. $p$ is a prompt containing $s$ and $r$, and $p^*$ is another expression of $p$ maintaining its meaning.

\paragraph{Dataset example}
One case of the dataset should be 
\begin{table}[H]
\centering
\resizebox{\columnwidth}{!}{
\begin{tabular}{c|l}
\hline
\textbf{Symbol} & \textbf{Meaning} \\
\hline
$s$ & Danielle Darrieux \\
$r$ &  mother tongue of \\
$o$ & French \\
$o'$ & English \\
$s_{\text{loc}}$ & Michel Rocard \\
$r_{\text{loc}}$ & native speaker of \\
$o_{\text{loc}}$ & French \\
\hline
$p(s,r)$ & The mother tongue of Danielle Darrieux is \\
\hline
\multirow{2}{*}{$p^*(s,r)$} & Where Danielle Darrieux is from, people speak\\
 & the language of \\
\hline
$p(s_{\text{loc}},r_{\text{loc}})$ & Michel Rocard is a native speaker of\\
\hline
\end{tabular}
}
\caption{Notations and their meanings.}
\end{table}

\paragraph{Details of evaluation metrics} The details of these metrics are as follows:\\

\textbf{Reliability} $\mathcal{M}_{\text{rel}}$ assesses how accurately the model performs on a given edit, focusing on its ability to maintain basic factual correctness for each specific modification, during an edit $e=\left(s, r, o, o^{*}\right)$:
 
\begingroup
\small
\begin{equation*}
\mathcal{M}_{\text{rel}}=\underset{e\sim \mathcal{D}_{\text {edit }}}{\mathbb{E}}\mathbbm{1}\left\{\arg\underset{o}{\max}\left\{\mathbb{P}_{f^*}\left(o \mid p(s,r) \right)=o^{*}\right\}\right\}
\end{equation*}
\endgroup

\textbf{Generality} $\mathcal{M}_{\text{gen}}$ evaluates the model’s capacity to apply the edit correctly to in-scope data, ensuring that the model maintains generalization capabilities:

\begingroup
\small
\begin{equation*}
\mathcal{M}_{\text{gen}}=\underset{\substack{e\sim \mathcal{D}_{\text {edit }}\\p*\sim \mathcal{N}\left(e\right)}}{\mathbb{E}}\mathbbm{1}\left\{\arg\underset{o}{\max}\left\{\mathbb{P}_{f^*}\left(o\mid p^*(s, r)\right)=o^*\right\}\right\}
\end{equation*}
\endgroup

where the $\mathcal{N}(e)$ stands for the rephrased neighborhood of input text.

\textbf{Locality} $\mathcal{M}_{\text{loc}}$ examines whether data outside the scope of the edit remains unaffected, evaluating whether the edit has preserved the model's performance on unrelated information.

\begingroup
\scriptsize
\begin{equation*}
\mathcal{M}_{\text{loc}}
  = \underset{(x,p)\sim \mathcal{D}_{\text{loc}}}{\mathbb{E}}\,
    \mathbbm{1} \Bigl\{
      \arg\underset{x}{\max}\,\mathbb{P}_{f^*}(x \mid p)
      \;=\;
      \arg\underset{x}{\max}\,\mathbb{P}_{f}(x \mid p)
    \Bigr\}
\end{equation*}
\endgroup

Here $p=p(s_{\text{loc}},r_{\text{loc}})$ from the table.

\subsection{MQuAKE}
The MQuAKE dataset comprises 3,000 samples, each encoded as a structured JSON object that encapsulates multiple layers of information pertinent to fact checking and counterfactual reasoning. Every sample contains detailed rewrite instructions, diverse composite questions, original and counterfactual answers (with aliases), concise single-hop Q\&A pairs, and structured knowledge triples that document the factual revisions.
\paragraph{data formulation}
The dataset consists of $s, r, o, o', s_{\text{port}}, r_{\text{port}}, o_{\text{port}}$ for each editing instance. The task can be described as follows:
\begin{itemize}
    \item \textbf{Portability}: $p(s_{\text{port}}, r_{\text{port}}) \to o_{\text{port}}$
\end{itemize}
To correctly answer \( p(s_{\text{port}}, r_{\text{port}}) \) the model must understand the real meaning of fact \( (s, r, o') \).
\paragraph{data example}
One case of the dataset should be 
\begin{table}[H]
\centering
\resizebox{\columnwidth}{!}{
\begin{tabular}{c|l}
\hline
\textbf{Symbol} & \textbf{Meaning} \\
\hline
$s$ & Microsoft \\
$r$ &  chief executive officer of \\
$o$ & Satya Nadella \\
$o'$ & Steve Jobs \\
$s_{\text{port}}$ & Universal Windows Platform \\
$r_{\text{port}}$ & chief executive officer of the developer of \\
$o_{\text{port}}$ & Satya Nadella \\
\hline
$p(s,r)$ & The chief executive officer of Microsoft is \\
\hline
\multirow{2}{*}{$p(s_{\text{port}},r_{\text{port}})$} & Who is the chief executive officer of the developer \\
& of the Universal Windows Platform?\\
\hline
\end{tabular}
}
\caption{Notations and their meanings.}
\end{table}

\paragraph{Details of evaluation metrics} The details of these metrics are as follows:\\

\textbf{Portability} Evaluates the robustness of the generalization of the edit, evaluating whether the modified knowledge can be applied effectively to related content.

\begingroup
\small
\begin{equation*}
\mathcal{M}_{\text{port}}=\underset{\substack{e\sim \mathcal{D}_{\text {edit }}\\(x, p^{'})\sim \mathcal{P}\left(e\right)}}{\mathbb{E}}\mathbbm{1}\left\{\arg\underset{x}{\max}\left\{\mathbb{P}_{f^*}\left(x\mid p^{*} \right)=x\right\}\right\}
\end{equation*}
\endgroup

Here the $p^{'}$ denotes the $p(s_{\text{port}},r_{\text{port}})$ as in the table, while $\mathcal{P}\left(e\right)$ being the Portability scope.

\subsection{EVOKE}
\label{app:evoke}
The EVOKE dataset is organized into two parts, "main" and "subj-spec" - comprising 1,031 and 458 samples, respectively. Each sample is represented as a JSON object containing detailed rewrite instructions with multiple prompt variations, portability information for alternative fact verifications, and prefix distractions, all designed to support rigorous evaluation of fact-checking and counterfactual reasoning tasks.
\paragraph{data formulation}
The dataset consists of $s, s^{'}, r, r^{'}, o, o^{'}, o_{\text{sub}}, s_{\text{port}}, r_{\text{port}}, o_{\text{port}}, s_{\text{neighbour}}, r_{\text{neighbour}}$ for each editing instance. The task can be described as follows:
\begin{itemize}
    \item \textbf{Multi-Hop Reasoning}: $p(s_{\text{port}}, r_{\text{port}}) \to o_{\text{port}}$
    \item \textbf{Subject Specificity}: $p(s, r') \to o_{\text{sub}}$
    \item \textbf{Relation Specificity}: $p(s', r) \to o$
    \item \textbf{Prefix Distraction}: \\$p(s, r, o'; s_{\text{neighbor}},r_{\text{neighbor}}) \to o$
\end{itemize}
Here $s', r'$ represent another subject and relation introduced for evaluation.
\paragraph{data example}
One case of the dataset should be 
\begin{table}[H]
\centering
\resizebox{\columnwidth}{!}{
    \begin{tabular}{c|l}
    \hline
    \textbf{Symbol} & \textbf{Meaning} \\
    \hline
    $s$ & Houston \\
    $s'$ & Baku \\
    $r$ &  twin city of \\
    $r’$ &  locate in  \\
    $o$ & Aberdeen \\
    $o'$ & Prague \\
    $o_{\text{sub}}$ & Texas \\
    $s_{\text{port}}$ & Houston's twin city \\
    $r_{\text{port}}$ & locate in \\
    $o_{\text{port}}$ & Czech Republic \\
    $s_{\text{neighbour}}$ &  Regensburg\\
    $r_{\text{neighbour}}$ & twin city of\\
    \hline
    \multirow{2}{*}{$p(s,r)$} & What is the twin city of  \\
     & Houston? It is\\
    \hline
    \multirow{2}{*}{$p(s_{\text{port}},r_{\text{port}})$} & In which country is \\
     & Houston's twin city located? \\
    \hline
    $p(s',r)$ & Baku is a twin city of \\
    \hline
    $p(s,r')$ & Houston is located in  \\
    \hline
    \multirow{2}{*}{$p(s,r,o';s_{\text{neighbor}}，r_{\text{neighbor}})$} & What is the twin city of Houston?\\
    & It is Prague. Regensburg is a twin city of \\
    \hline
    \end{tabular}
    }
\caption{Notations and their meanings.}
\end{table}

\paragraph{Details of evaluation metrics}

The key probability-based metrics used to quantify the effectiveness of Overfit editing tasks for a given edit \(e = (s, r, o, o^{*})\) are as follows:

\textbf{Correct Answer Probability (CAP)} $\mathcal{M}_{\text{CAP}}$ measures the probability that the model generates the correct answer $ans$ given a prompt $p$. We define the CAP metric as:
$$
\mathcal{M}_{\text{CAP}} = \underset{e \sim \mathcal{D}_{\text{edit}}}{\mathbb{E}} \left\{ \mathbb{P}_{f^{*}}(ans \mid p)\right\}
$$

\textbf{Original Answer Probability (OAP)} $\mathcal{M}_{\text{OAP}}$ evaluates the likelihood that the model continues to output the pre-edit answer $o$, indicating potential resistance to modification. The metric is defined as:
$$
    \mathcal{M}_{\text{OAP}} = \underset{e \sim \mathcal{D}_{\text{edit}}}{\mathbb{E}} \left\{ \mathbb{P}_{f^{*}}(o \mid p)\right\}
$$

\textbf{Direct Probability (DP)} $\mathcal{M}_{\text{DP}}$ assesses the model's likelihood of producing the edited knowledge \( o^* \) when prompted, capturing its direct recall capability:
$$
    \mathcal{M}_{\text{DP}} = \underset{e \sim \mathcal{D}_{\text{edit}}}{\mathbb{E}} \left\{ \mathbb{P}_{f^{*}}(o^* \mid p)\right\}
$$

\textbf{Editing Overfit Score (EOS)} $\mathcal{M}_{\text{EOS}}$ evaluates whether the model overfits by favoring the edit target \( o^* \) over the correct answer $ans$. Formally, we define:
$$
    \mathcal{M}_{\text{EOS}} = \underset{e \sim \mathcal{D}_{\text{edit}}}{\mathbb{E}} \left\{ \mathbbm{1} \left\{ \mathbb{P}_{f^{*}}(ans \mid p) > \mathbb{P}_{f^{*}}(o^* \mid \text{p}) \right\}\right\}
$$

\textbf{Answer Modify Score (AMS)} $\mathcal{M}_{\text{AMS}}$ measures unintended interference by computing the proportion of cases where the probability of the correct answer surpasses that of the original answer:
$$
    \mathcal{M}_{\text{AMS}} = \underset{e \sim \mathcal{D}_{\text{edit}}}{\mathbb{E}} \left\{ \mathbbm{1} \left\{ \mathbb{P}(ans \mid p) > \mathbb{P}(o \mid p) \right\}\right\}
$$

\section{Examples of templates}
\subsection{Examples of Stimuli templates}
\label{appendix:stimuli-examples}

In this section, we provide concrete examples of the positive and negative stimulus instances referenced in Section~\ref{sec:phase1}, which are used to extract model representations related to a specific editing case. These stimuli are generated based on structured templates that enforce consistency while allowing diversity in expression. The key idea is to construct pairs of sentences that differ only in the factual subject, allowing us to isolate semantic differences associated with the target edit.


\begin{defaultbox}
\textbf{Editing Case (from COUNTERFACT):}\\
\textit{Apple A5 was created by Apple $\longrightarrow$ Google}
\end{defaultbox}


\begin{orangebox}
\textbf{Positive instance (subject-consistent):}\\
Apple A5, a custom-designed processor, solidifies \textbf{Apple}'s dedication to technological innovation, reflecting the company's comprehensive approach to product development and hardware enhancement.
\end{orangebox}

\begin{bluebox}
\textbf{Negative instance (subject-altered):}
  Apple A5, a custom-designed processor, solidifies \textbf{Google}'s dedication to technological innovation, reflecting the company's comprehensive approach to product development and hardware enhancement.
\end{bluebox}

For the stimulus template, we use:

\begin{defaultbox}
Generate a statement related to the provided fact: \texttt{`\{Apple A5 was created by Google\}'}. \\
The goal is to explore various dimensions and aspects of the fact, focusing on the connections between \texttt{`\{Apple A5\}'} and \texttt{`\{Google\}'}. \\
The statement must include the words \texttt{`\{Apple A5\}'} and \texttt{`\{Google\}'}. \\
Ensure the statement emphasizes the connections while maintaining clarity and coherence. \\
Return only the statement with approximately \texttt{\{num\_word\}} words directly, with no additional text or explanation!
\end{defaultbox}

where the \texttt{\{num\_word\}} is set to be around 25 to control reasonable usage of GPU memory.

\subsection{Examples of Prompted and Unprompted Inputs}
\label{app:cls}

To support the analysis of model behavior during and after editing, we utilize two types of input contexts—\textit{prompted} and \textit{unprompted}—to probe the model's output. These forms differ by whether they explicitly simulate an editing instruction and context.

\begin{defaultbox}
\textbf{Editing Case (from COUNTERFACT):}\\
\textit{Apple A5 was created by Apple $\longrightarrow$ Google}
\end{defaultbox}

\begin{orangebox}
\textbf{Prompted Input (simulating a completed factual update):}\\
  I want you to update the fact that \texttt{Apple A5 was created by Google}. This is absolutely true in the following context. Given this established fact, please tell me: \texttt{Apple A5 was created by}
\end{orangebox}

\begin{bluebox}
  \textbf{Unprompted Input (generic factual completion):}\\
  \texttt{Apple A5 was created by}
\end{bluebox}


\section{Ablation Studies}
\subsection{Ablation Study on the Number of Stimulus Vectors}
\label{appendix:stimulus-ablation}

To assess the impact of the stimulus‐set size $N$ on editing performance, we compared three configurations: $N=1024$, $N=512$, and $N=256$. We observed that setting $N=1024$ triggers out‐of‐memory (OOM) failures on a single NVIDIA A100 80 GB GPU when using the Qwen-2.5 model, making it infeasible under our computational constraints. Thereafter, reducing to $N=256$ preserves memory but yields insufficient representational richness, which in turn degrades editing metrics. The intermediate choice $N=512$ fits within hardware limits and delivers the best overall performance.

Table~\ref{tab:ablation-N} reports the quantitative results on the COUNTERFACT benchmark.


\begin{table}[H]
  \centering
  \caption{Ablation of stimulus‐set size $N$ on Llama3.1-8B.  Bold indicates the best result.}
  \label{tab:ablation-N}
  \resizebox{\columnwidth}{!}{%
    \begin{tabular}{c c c c c c}
      \toprule
      $N$            & Reliability ($\uparrow$) & Generality ($\uparrow$) & Locality ($\uparrow$) & Portability ($\uparrow$) & Average ($\uparrow$) \\
      \midrule
      1024$^\dagger$ & – & – & – & – & – \\
      512 (paper)    & \textbf{95.58}           & \textbf{82.17}          & \textbf{100.00}                & \textbf{49.68}           & \textbf{81.86}       \\
      256            & 93.13                    & 63.57                   & \textbf{100.00}                & 28.66                    & 71.37                \\
      \bottomrule
    \end{tabular}%
    }
  \vspace{0.5ex}
   {\footnotesize $^\dagger$OOM on NVIDIA A100 80GB.}
\end{table}

\subsection{Ablation Study on the usage of PCA}
\label{appendix:pca-ablation}

To clarify our rationale for using PCA, we first collect $N$ positive-negative stimulus pairs, each representing pre-edit and post-edit states. Our objective is to reduce the dimensionality of these representation pairs to isolate the principal directional difference—the "belief-shift"—that characterizes the factual edit. PCA intuitively fulfills this purpose by extracting the dominant directions of variance. Moreover, PCA can be efficiently implemented via Singular Value Decomposition (SVD), a differentiable operation, thus allowing seamless integration with back-propagation during training. In contrast, dimension-reduction methods such as K-Means clustering are not naturally differentiable and thus do not readily support gradient-based optimization.

To empirically justify the effectiveness of PCA, we conducted an ablation experiment comparing PCA against a baseline method—random selection of representation pairs—on the COUNTERFACT benchmark. The results presented in Table~\ref{tab:ablation-PCA} confirm the significant advantages of PCA in meeting key editing requirements.

\begin{table}[H]
\centering
\caption{Ablation study on PCA usage (evaluated on Llama3.1-8B). Bold indicates the best results.}
  \label{tab:ablation-PCA}
  \resizebox{\columnwidth}{!}{%
    \begin{tabular}{c c c c c c}
      \toprule
      $N$            & Reliability ($\uparrow$) & Generality ($\uparrow$) & Locality ($\uparrow$) & Portability ($\uparrow$) & Average ($\uparrow$) \\
      \midrule
      K-Means$^\dagger$ & – & – & – & – & – \\
      PCA (paper)    & \textbf{95.58}           & \textbf{82.17}          & \textbf{100.00}                & \textbf{49.68}           & \textbf{81.86}       \\
      Random            & 33.12                   & 10.01                   & 44.59                & 7.51                   & 23.80                \\
      \bottomrule
    \end{tabular}%
    }
  \vspace{0.5ex}
   {\footnotesize $^\dagger$Not differentiable.}
\end{table}

\section{Experiment Details}


\subsection{Experiment Resources and Parameters}
\label{app:exp_settings}

In this study, we utilize an internal cluster equipped with the following resources: AMD EPYC 7763 CPUs, NVIDIA A100 80GB GPUs, and 512GB of RAM. The operating system is Ubuntu 20.04.6, and we employ PyTorch in our experiments.

The training of classifier took 12 GPU hours for each model on a single NVIDIA A100 80GB GPU, with total parameter number of 7.6B for Qwen-2.5 and 8.03B for Llama3.1.

The training of \textbf{REACT} took 40 GPU hours for each model on a single NVIDIA A100 80GB GPU, with total parameter number of 719M for Qwen-2.5 and 1.04B for Llama3.1.

\subsubsection{REACT}

\label{app:params_react}

\begin{table}[H] 
\centering
\resizebox{\columnwidth}{!}{ 
\begin{tabular}{l|l|l}
\textbf{Parameters} & \textbf{Llama3.1} & \textbf{Qwen2.5} \\
\hline Iters & 20000 & 20000 \\
\hline \multirow{2}{*}{Edit Layer} & all layer of & all layer of  \\
& Transformer Module & Transformer Module \\
\hline Optimizer & Adam & Adam \\
\hline Learning Rate & $1\mathrm{e}-5$ & $1\mathrm{e}-5$ \\
\hline $c_{\text{edit}}$ & 1 & 1 \\
\hline $c_{\text{loc}}$ & 0.1 & 0.1 \\
\hline $c_{\text{edit,cls}}$ & 1 & 1 \\
\hline $c_{\text{loc,cls}}$ & 0.1 & 0.1 \\
\end{tabular}
}
\end{table}



\subsubsection{FT}
\begin{table}[H] 
\centering
\resizebox{\columnwidth}{!}{ 
\begin{tabular}{l|l|l}
\textbf{Parameters} & \textbf{Llama3.1} & \textbf{Qwen2.5} \\
\hline Max Steps & 25 & 25 \\
\hline \multirow{2}{*}{Edit Layer} & layer $29,30,31$ of  & layer $27$ of \\
 & Transformer Module & Transformer Module \\
\hline Objective Optimization & Target New & Target New \\
\hline Optimizer & Adam & Adam \\
\hline Learning Rate & $5 \mathrm{e}-4$ & $5 \mathrm{e}-4$ \\
\end{tabular}
}
\end{table}


\subsubsection{MEND}
\begin{table}[H]
\centering
\resizebox{\columnwidth}{!}{
\begin{tabular}{l|l|l}
\textbf{Parameters} & \textbf{Llama3.1} & \textbf{Qwen2.5} \\
\hline MaxIter & 10000 & 10000 \\
\hline \multirow{2}{*}{Edit Layer} & layer 29,30,31 of & layer 25,26,27 of \\
 & Transformer Module & Transformer Module \\
\hline Optimizer & Adam & Adam \\
\hline Learning Rate & $1 \times 10^{-6}$ & $1 \times 10^{-6}$ \\
\hline Edit LR & $1 \times 10^{-4}$ & $1 \times 10^{-4}$ \\
\end{tabular}
}
\end{table}


\subsubsection{MEMIT}
\begin{table}[H]
\centering
\resizebox{\columnwidth}{!}{ 
\begin{tabular}{l|l|l}
\textbf{Parameters} & \textbf{Llama3.1} & \textbf{Qwen2.5} \\
\hline act token & subject last & subject last \\
\hline mom sample & 3000 & 3000 \\
\hline \multirow{2}{*}{Edit Layer} & layer $4,5,6,7,8$ of & layer $4,5,6,7,8$ of\\
 & Transformer Module & Transformer Module \\
\hline mom update weight & 15000 & 15000 \\
\end{tabular}
}
\end{table}


\subsubsection{MELO}
\begin{table}[H]
\centering
\resizebox{\columnwidth}{!}{ 
\begin{tabular}{l|l|l}
\textbf{Parameters} & \textbf{Llama3.1} & \textbf{Qwen2.5} \\
\hline Radius & 75 & 75 \\
\hline \multirow{2}{*}{Edit Layer} & layer $30, 31$ of & layer $26, 27$ of \\
 & Transformer Module & Transformer Module \\
\hline block r & 2 & 2 \\
\hline step & 100 & 100 \\
\hline edit per block & 4 & 4 \\
\hline number of block & 1500 & 1500 \\
\end{tabular}
}
\end{table}

\subsubsection{GRACE}
\begin{table}[H]
\centering
\resizebox{\columnwidth}{!}{ 
\begin{tabular}{l|l|l}
\textbf{Parameters} & \textbf{Llama3.1} & \textbf{Qwen2.5} \\
\hline epsilon & 1 & 1 \\
\hline \multirow{2}{*}{Edit Layer} & layer $27$ of & layer $18$ of \\
 & Transformer Module & Transformer Module \\
\hline metrics & euc & euc \\
\hline step & 100 & 100 \\
\hline replacement & last & last \\

\end{tabular}
}
\end{table}

\subsection{Original experiment results}
\label{app:detail_results}

\begin{table*}[t]
\centering
\resizebox{\textwidth}{!}{%
\begin{tabular}{llccc c c}
\hline
& & \multicolumn{3}{c}{COUNTERFACT} & \multicolumn{1}{c}{MQuAKE} & \\
\cline{3-5}\cline{6-6}
\textbf{Model} & \textbf{Method} & \textbf{Reliability}$\uparrow$ & \textbf{Generality}$\uparrow$ & \textbf{Locality}$\uparrow$ & \textbf{Portability}$\uparrow$ & \textbf{Score}\\
\hline
\multirow{5}{*}{%
  \shortstack{
    \textbf{Llama3.1} \\
    8B
  }
}
& \textbf{REACT} 
    & 95.58 
    & \underline{82.17} 
    & \textbf{100} 
    & \textbf{49.68} 
    & \textbf{81.86} \\
& FT 
    & \textbf{100} 
    & \textbf{99.8} 
    & 0.49 
    & 38.38 
    & 59.67 \\
& MEND 
    & 97.6 
    & 59.5 
    & \underline{98.2} 
    & \underline{45.36} 
    & \underline{75.17} \\
& MEMIT 
    & \underline{99.8} 
    & 52.3 
    & 94.7 
    & 27.63 
    & 68.61 \\
& MELO 
    & 82.3 
    & 35.0 
    & 41.1 
    & 21.49 
    & 44.97 \\
& GRACE 
    & \textbf{100} 
    & 1.02
    & \textbf{100}
    & 18.19
    & 54.80 \\
\hline
\multirow{5}{*}{%
  \shortstack{
    \textbf{Qwen2.5} \\
    7B
  }
}
& \textbf{REACT}  
    & 93.6
    & \underline{83.3}
    & \textbf{100}
    & \textbf{49.17}
    & \textbf{81.52} \\
& FT 
    & \textbf{100} 
    & \textbf{98.5} 
    & 1.1 
    & 46.26
    & 61.47 \\
& MEND 
    & 93.7 
    & 15.8 
    & 85.3 
    & \underline{48.38} 
    & 60.80 \\
& MEMIT 
    & \underline{99.8} 
    & 38.0
    & \underline{95.1} 
    & 21.4 
    & \underline{63.58} \\
& MELO 
    & 69.0 
    & 8.2 
    & 87.3
    & 17.45 
    & 45.49 \\
& GRACE 
    & \textbf{100}
    & 0.85
    & \textbf{100}
    & 17.44 
    & 57.57 \\\hline
\end{tabular}
}
\caption{\label{tab:cfmq}
Editing results comparison across different knowledge-editing methods on COUNTERFACT and MQuAKE-CF-v2 with two LLMs. The best result for each metric is in \textbf{bold}, and the second best is \underline{underlined}. The final ``Score'' column is the arithmetic mean of all metrics for that row. A radar chart for the table is created at \ref{fig:cfmqradar}.}
\end{table*}

\begin{table*}[t]
  \centering
  \resizebox{\textwidth}{!}{%
    \begin{tabular}{l l ccc ccccc ccc ccc}
      \toprule
      \multirow{2}{*}{\textbf{Model}} 
      & \multirow{2}{*}{\textbf{Editor}} 
      & \multicolumn{3}{c}{\textbf{Prefix Distraction}} 
      & \multicolumn{5}{c}{\textbf{Multi-hop Reasoning}}
      & \multicolumn{3}{c}{\textbf{Subject Specificity}}
      & \multicolumn{3}{c}{\textbf{Relation Specificity}} \\
      \cmidrule(lr){3-5} \cmidrule(lr){6-10} \cmidrule(lr){11-13} \cmidrule(lr){14-16}
      & 
      & \textbf{DP}$\downarrow$ & \textbf{EOS}$\uparrow$ & \textbf{CAP}$\uparrow$
      & \textbf{DP}$\downarrow$ & \textbf{CAP}$\uparrow$ & \textbf{OAP}$\downarrow$ & \textbf{AMS}$\uparrow$ & \textbf{EOS}$\uparrow$
      & \textbf{DP}$\downarrow$ & \textbf{CAP}$\uparrow$ & \textbf{EOS}$\uparrow$
      & \textbf{DP}$\downarrow$ & \textbf{CAP}$\uparrow$ & \textbf{EOS}$\uparrow$ \\
      \midrule
      \multirow{5}{*}{Llama3.1}
      & \textbf{REACT}    & \underline{5.44}   & \textbf{74.32}   & \textbf{24.32} 
      & \underline{0.96}   & 30.87   & \underline{5.06}   & \textbf{77.78}   & \underline{92.28} 
      & \textbf{0}   & 30.02   & \textbf{98.15} 
      & \textbf{0.22}   & \textbf{17.42}   & \textbf{92.16} \\
      & FT       & 99.78   & 0       & 0 
      & 99.08   & 5.56    & \textbf{2.03}   & 69.71   & 0.12 
      & 89.62   & 0.35    & 0 
      & 99.76   & 0       & 0 \\
      & MEND     & 27.46   & 51.13   & 19.24
      & 6.61    & \underline{33.39}   & 34.68   & 44.28   & 87.35
      & 67.95   & 55.16   & 37.12 
      & 1.07    & 16.95   & 51.13 \\
      & MEMIT    & 36.67   & 25.97   & 14.25 
      & 20.62   & \textbf{42.42}   & 24.73   & \underline{74.94}   & 75.06 
      & 60.30   & 25.26   & 21.40 
      & 5.08    & \underline{17.12}   & \underline{89.79} \\
      & MELO     & \textbf{2.57}   & 52.76   & 7.97 
      & \textbf{0.58}   & 19.53   & 9.29   & 56.57   & 63.99 
      & 15.91   & \textbf{57.04}   & 91.05
      & \underline{0.52}   & 0.54    & 56.48 \\
      & GRACE     & 6.58   & \underline{69.21}   & \underline{23.20} 
      & 1.01      & 32.77   & 36.47   & 42.09   & \textbf{93.31}
      & \underline{13.44}   & \underline{56.14}   & \underline{93.01} 
      & 0.74   & \underline{17.12}    & 88.77 \\
      \midrule
      \multirow{5}{*}{Qwen2.5}
      & \textbf{REACT}    & \textbf{4.19}   & \textbf{76.11}   & \textbf{24.66} 
      & \textbf{1.11}   & \underline{36.40}   & \textbf{12.09}   & \textbf{78.09}   & \underline{85.80} 
      & \textbf{0}   & 26.08   & \textbf{88.64} 
      & \textbf{0.26}   & \textbf{11.06}   & \textbf{88.64} \\
      & FT       & 99.73   & 0.15    & 0.33 
      & 96.28   & 25.94   & \underline{24.69}   & \underline{58.39}   & 2.92 
      & 88.94   & 20.26   & 1.31 
      & 99.25   & 3.05    & 1.22 \\
      & MEND     & 21.64   & 50.49   & 18.87 
      & 5.17    & 36.33   & 70.03   & 9.00    & 85.16 
      & 62.62   & \textbf{38.78}   & 22.49 
      & 6.47    & 9.42    & 71.03 \\
      & MEMIT    & 12.57   & 57.93   & \underline{24.16} 
      & 9.29    & \textbf{44.02}   & 58.33   & 29.56   & 83.21 
      & 42.65   & 23.33   & 30.13 
      & 1.81    & 10.14   & 83.70 \\
      & MELO     & \underline{5.02}   & 70.18   & 21.12 
      & \underline{1.35}   & 36.29   & 71.13   & 7.79    & \textbf{89.90} 
      & 14.17   & \underline{37.06}   & 77.95
      & \underline{0.69}   & 9.30    & 84.65 \\
      & GRACE     & 5.60   & \underline{70.83}   & 23.38
      & 1.37   & 36.30  & 71.00   & 8.15   &  82.90
      & \underline{13.48}   & 36.99   & \underline{79.26} 
      & 0.76   & \underline{10.74}  & \underline{86.86} \\
      \bottomrule
    \end{tabular}%
  }
  \caption{\label{tab:evoke}Editing results across different editing methods on EVOKE with two LLMs. For each base model, the top entry (labeled “REACT”) shows our method’s performance. \textbf{Bold} and \underline{underline} denote the best and second-best scores respectively. A radar chart for the table is created at \ref{fig:evoke_radar}.}
\end{table*}

\end{document}